\pdfoutput=1
\documentclass[11pt]{article}

\usepackage[]{emnlp2021}

\usepackage{times}
\usepackage{latexsym}
\usepackage{amssymb}
\usepackage{amsmath}
\usepackage{courier}
\usepackage{multirow}
\usepackage{graphicx}
\usepackage{booktabs}
\usepackage{hyperref}
\usepackage{bbding}
\usepackage{graphicx}
\usepackage{subfigure}

\usepackage[T1]{fontenc}
\usepackage[utf8]{inputenc}

\usepackage{microtype}

\title{Increasing Visual Awareness in Multimodal Neural Machine Translation \\from an Information Theoretic Perspective}

\begin{document}

% \author{ 
% Baijun Ji$^\S$,  Tong Zhang$^\S$, Yicheng Zou${^\ddag}$,  Bojie Hu$^\S$\thanks{\, Corresponding author}
% \, and  Si Shen$^{\dag}^\ast$\\
%   $^\S$ Tencent TEG AI\\
%   $^\ddag$ School of Computer Science, Fudan University \\
%   $^\dag$ Research Base on Interdisciplinary Terminology and Translation, Nanjing University \\
%   $^\S$\{baijunji,zatozhang, bojiehu\}@tencent.com \quad
%   $^\ddag$  yczou18@fudan.edu.cn  
%  \
% }

\author{ 
Baijun Ji$^1$,  Tong Zhang$^1$, Yicheng Zou${^2}$,  Bojie Hu$^1$\thanks{\, Corresponding author} , Si Shen$^{3}$\\
  $^1$ Tencent Minority-Mandarin Translation, Beijing, China \\
  $^2$ School of Computer Science, Fudan University \\
  $^3$ Research Base on Interdisciplinary Terminology and Translation, Nanjing University \\
  $^1$\{baijunji,zatozhang, bojiehu\}@tencent.com \quad
   $^2$  yczou18@fudan.edu.cn  
 \
}

\maketitle
\begin{abstract}
  % \begin{quote}
    Multimodal machine translation (MMT) aims to improve translation quality by equipping the source sentence with its corresponding image. Despite the promising performance, MMT models still suffer the problem of input degradation: models focus more on textual information while visual information is generally overlooked. In this paper, we endeavor to improve MMT performance by increasing visual awareness from an information theoretic perspective. In detail, we decompose the informative visual signals into two parts: source-specific information and target-specific information. We use mutual information to quantify them and propose two methods for  objective optimization to better leverage visual signals. Experiments on two datasets demonstrate that our approach can effectively enhance the visual awareness of MMT model and achieve superior results against strong baselines.
\end{abstract}

\section{Introduction}
Multimodal machine translation (MMT) typically improves text translation quality by introducing an extra visual modality input. This task hypothesizes that the corresponding visual context is helpful for disambiguation and omission completion when the source sentence is ambiguous or even incorrect. Compared to the purely text-based machine translation, the critical point of MMT is to find out an effective approach to integrating images into a MT model that makes the most of visual information.

Most of the existing studies have dedicated their efforts to the way of extracting multi-granularity visual features for integration \cite{Calixto2017IncorporatingGV, Delbrouck2017AnES,Ive2019DistillingTW,Zhao2022RegionattentiveMN,Li2022OnVF,Fang2022NeuralMT} or designing model architectures for better message passing across various modalities \cite{Calixto2017DoublyAttentiveDF, Calixto2019LatentVM, Yao2020MultimodalTF, Yin2020ANG,Lin2020DynamicCC, Liu2021GumbelAttentionFM}, which achieve promising performances in most of the multimodal scenarios.

% However, some doubt has been cast upon the effectiveness of visual modality in MMT \cite{Grnroos2018TheMS, Elliott2018AdversarialEO}. Elliott
% \shortcite{Elliott2018AdversarialEO} proposed an adversarial evaluation method to judge whether a MMT model is able to perceive visual context. They found that the performance of some public MMT models only dropped a little when given an incongruent image. The phenomenon suggests that MMT models are not sensitive to this perturbation, and visual context is actually underutilized.
% A more recent study \cite{Wu2021GoodFM} claimed that the gains from the multimodal
% signals over text-only counterparts are, in fact, due to the regularization effect.
% They only equipped the MMT model with an elementary gate fusion network, which can easily outperform the previous well-designed models. From this perspective, the usage of visual modality is still not fully explored in the most of previous works.

% In this paper, we argue that the visual modality remains an excellent potential to improve the translation performance as long as the MMT model is visual-aware enough.
% It is a common practice that MMT models employ maximum likelihood estimation (MLE) for training. Nevertheless, few approaches have introduced other constraints to better leverage visual context. As a result, MMT models tend to acquire knowledge from the textual modality even if text parts are ambiguous or incorrect, leading to the aggravation of the input degradation problem on visual signals.

Despite the success of these methods, some researchers reveal that visual modality is underutilized by the modern MMT models \cite{Grnroos2018TheMS, Elliott2018AdversarialEO}.
Elliott \shortcite{Elliott2018AdversarialEO} leverage an adversarial evaluation method to
appraise the awareness of visual contexts of MMT models. An interesting observation is that the disturbed image input generally shows no significant effects on the performance of public MMT models, indicating the insensitivity to the visual contexts.
Recently, \citet{Wu2021GoodFM} revisit the need for visual context in MMT and find that their strong MMT model tends to ignore the multimodal information. They suggest that 
the gains from the multimodal signals over text-only counterparts are, in fact, due to the regularization effect.

In practice, the existing approaches mainly focus on a better interaction method between textual- and visual- information. In training, they generally employ maximum likelihood estimation (MLE) to optimize the entire model. In most cases, the source sentence carries sufficient context for translation, leading to the neglect of visual modality under such training paradigm. However, the visual information is still crucial when the source sentence faces the problem of incorrection, ambiguity and gender-neutrality . Accordingly, we argue that the visual modality remains an excellent potential to facilitate translation, which inspire us to explore taking full advantage of the visual modality in MMT.

To this end, we propose increasing visual awareness from an information theoretic perspective. 
Specifically, we quantify the amount of visual information with an information-theoretic metric, namely Mutual Information (MI). The quantification is then divided into two parts: 1) the source-specific part, which means the information between source texts and images; 2) the target-specific part, which means the mutual information between target texts and images given source texts, namely conditional mutual information (CMI). For the source-specific part, we maximize the mutual information by resorting to a lower bound of MI. For the target-specific part, we propose a novel optimization method for maximizing the conditional mutual information (See details in \ref{sec:target-mi}). We evaluate our approach on two public datasets. The experimental results and in-depth analysis indicate that our method can effectively enhance the MMT model's sensitivity to visual context, which manifests that MMT has great potential when the utilization of visual context is adequate. 

To conclude, the main contributions of our paper are three-fold:
\begin{itemize}
  \item [1)] We propose a novel MI-based learning framework to improve multimodal machine translation by increasing visual awareness. This framework is interpretable and can quantify how much visual information is utilized.
 \item [2)] To optimize the mutual information objective, we divide the corresponding visual information into source/target-specific part. Furthermore, we propose a novel method to maximize the conditional mutual information in the target-specific part.
  \item [3)] Comprehensive experiments demonstrate that our method can effectively encourage MMT models to keep sensitive to visual context and significantly outperform strong baselines on two public MMT datasets.

\end{itemize}

  \section{Background}
  \subsection{Multimodal Machine Translation} 
  We start with the formulation of a regular MMT task. 
  Given a triplet dataset of $\{(x^{(i)},y^{(i)},z^{(i)})\}_1^N$, the problem naturally turns into the likelihood maximization:
  \begin{align}
      \mathcal{L}_{MMT}
      = -\frac{1}{N} \sum_{i=1}^{N} \log{p(y^{(i)}|x^{(i)},z^{(i)})},
  \end{align}
  where $z$ denotes the image itself and $x,y$ denote the description of the image in two different languages. Recently, MMT models are usually consisted of a Transformer-based encoder-decoder model and an additional image encoder. There exist several lines of works to integrate convolutional features into the NMT model. We adopt the global feature, which is in the form of a single vector of an image. This integration method has been proven beneficial to some extent in prior works.

\begin{figure}[tbp]
  % \centering % 图片居中
  \includegraphics[width = 7.5cm]{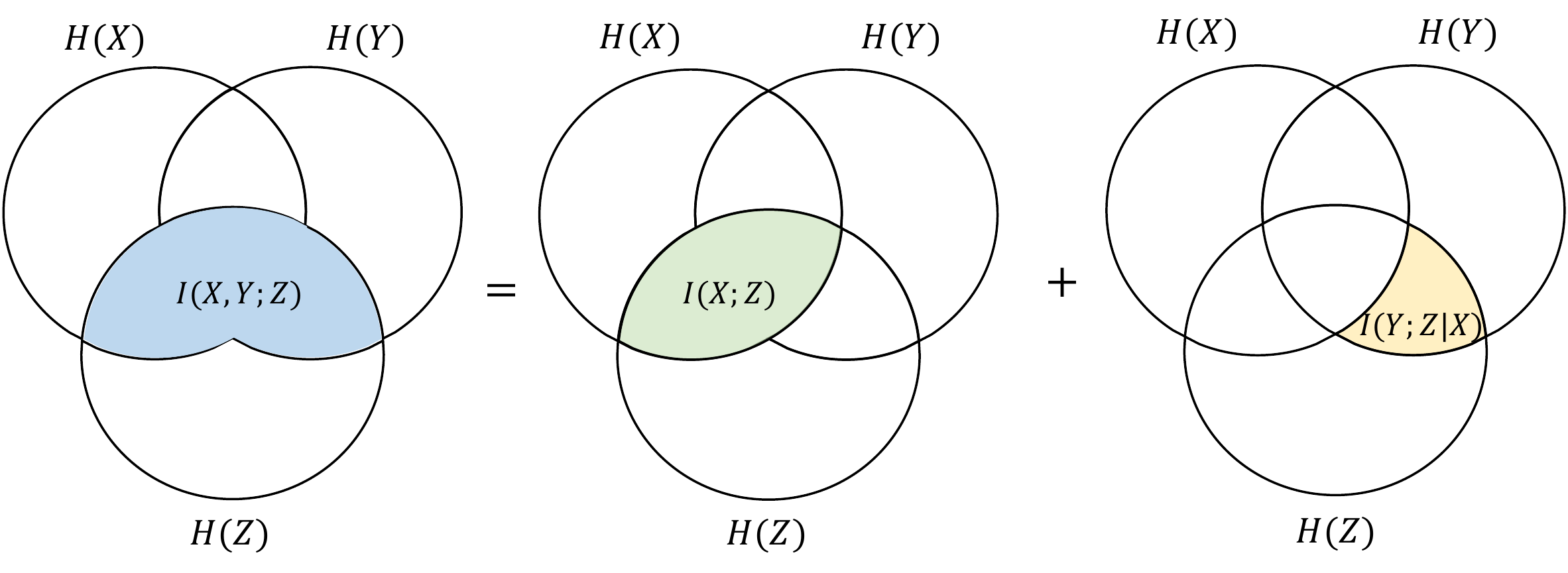}
  \caption{Chain rule for multual information. We decompose visual awareness into two parts: 1) mutual information between source text and image and 2) conditional mutual information between target text and image.}
  \label{fig:chain_rule}
\end{figure}

  \subsection{Mutual Information (MI)}
  Mutual Information (MI) relates two random variables $\mathrm{X}$ and $\mathrm{Y}$ and measures the amount of information obtained from $\mathrm{X}$ by observing the random variable $\mathrm{Y}$, which is defined as:
  \begin{align}
  I(\mathrm{X};\mathrm{Y})=H(\mathrm{X})-H(\mathrm{X}|\mathrm{Y}),
  \end{align}
  where $H(\cdot)$ denotes information entropy. Mutual information is symmetric and non-negative. High mutual information shows a large reduction of the random variable $X$'s uncertainty given another variable $Y$.

\section{Methodology}

\begin{figure}[tbp]
  % \centering % 图片居中
  \includegraphics[width = 7.5cm]{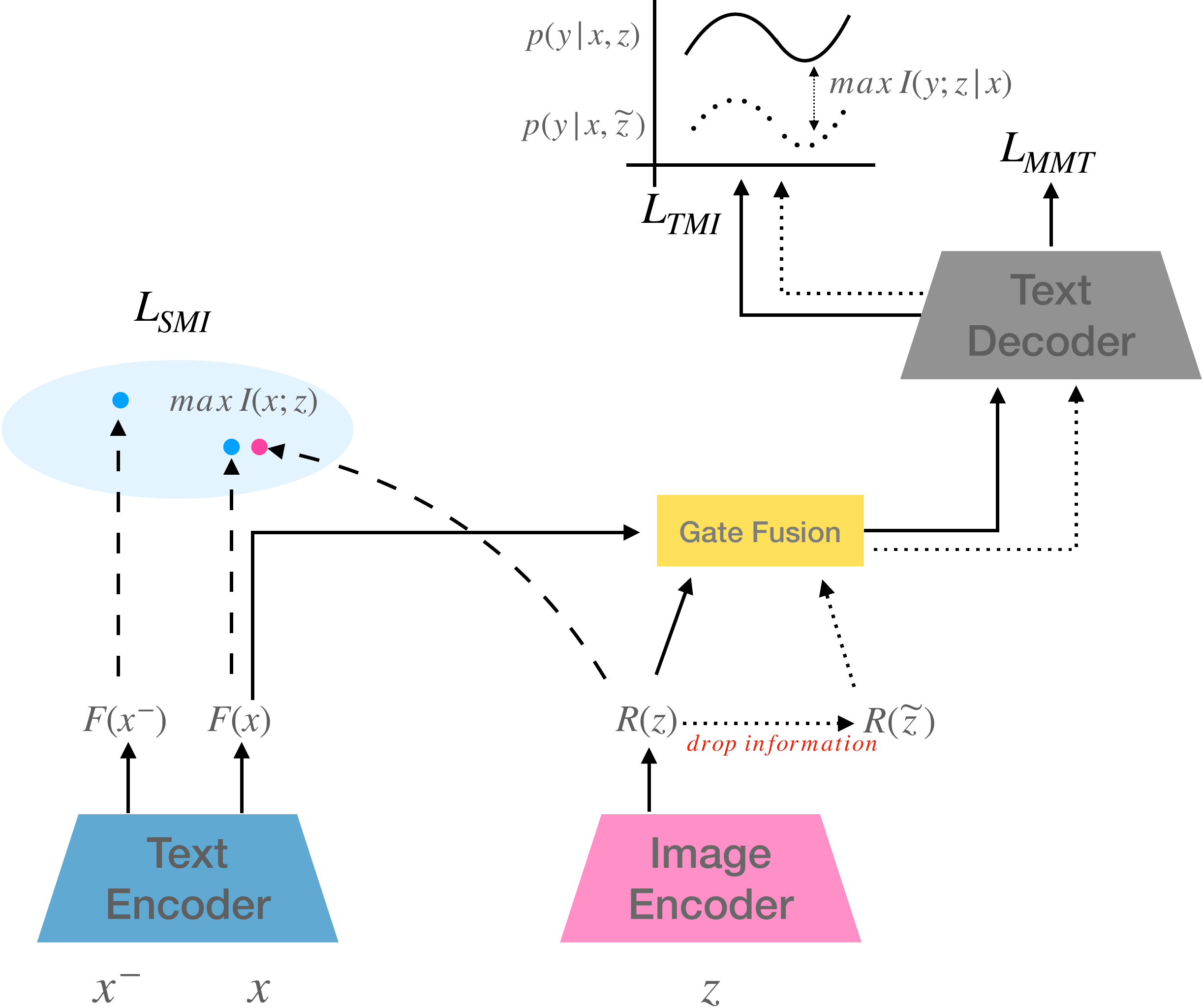}
  \caption{An overview of our method, where $x$ and $z$ 
  represent the source text and image, $x^-$ is a contrastive sample of $x$, and $\widetilde{z}$ is a deteriorated version of $z$ by removing visual information. $L_{SMI}$ aims to maximize MI between $x$ and $z$ to learn grounded representation. $L_{TMI}$ aims to maximize between $y$ and $z$ to strengthen the effect of visual context in generation. }
  \label{fig:model}
\end{figure}

In this section, we mathematically describe the basic notion of our proposed method from an information theoretic perspective. 
Let $\mathrm{X}$ denotes a random variable over source sentences, $\mathrm{Y}$ denotes a random variable over target sentences, and $\mathrm{Z}$ is a random variable over images. 
They are jointly distributed according to some true p.m.f. $p(x, y, z)$. We formulate the \textbf{Visual Awareness} of a MMT model as a joint mutual information $I(\mathrm{X},\mathrm{Y}; \mathrm{Z})$, which indicates how much the visual modality contributes to the overall system. A high joint mutual information suggests a strong sensitivity of the MMT model to visual information. However, it is intractable to directly maximize this joint mutual information in training. Thus, in this work we decompose $I(\mathrm{X},\mathrm{Y}; \mathrm{Z})$ into two parts by the chain rule of mutual information and optimize these two parts separately:
\begin{align}
    I(\mathrm{X},\mathrm{Y}; \mathrm{Z})= \underbrace{I(\mathrm{X};\mathrm{Z})}_{source-specific}+ 
      \underbrace{I(\mathrm{Y};\mathrm{Z}|\mathrm{X})}_{target-specific}.
\end{align}

As shown in Figure \ref{fig:chain_rule}, the green part $I(\mathrm{X};\mathrm{Z})$ is the source-specific mutual information (SMI), which indicates the mutual information between source texts and images; the yellow part $I(\mathrm{Y};\mathrm{Z}|\mathrm{X})$ is the target-specific conditional mutual information (TMI), which indicates the mutual information between target texts and images given source texts. Figure \ref{fig:model} illustrates the overview of our proposed method. In the following, we describe how to optimize the source-specific and target-specific mutual information in detail.

  % \subsection{$I(\mathrm{X};\mathrm{Z})$}
  \subsection{Source-Specific Mutual Information}
  Source-specific mutual information $I(\mathrm{X};\mathrm{Z})$ determines how much information is shared between the source texts and images. 
  Inspired by \citet{Oord2018RepresentationLW}, we minimize the $\textbf{InfoNCE}$ \cite{Logeswaran2018AnEF} loss $\mathcal{L}_{SMI}$ to maximize a lower bound on $I(\mathrm{X};\mathrm{Z})$. 
  \begin{align}
  I(\mathrm{X};\mathrm{Z}) \geq  -&\mathbb{E}_{p(x,z)} \biggr[
      \mathit{f}(\mathcal{F}(x), \mathcal{R}(z)) - \nonumber\\
      \mathbb{E}_{q(x^-)} \log &   
      \biggr[ \sum_{x^-}  \exp{\mathit{f}(\mathcal{F}(x^-), \mathcal{R}(z))} \biggr]
   \biggr]  \nonumber \\
    =& \mathcal{L}_{SMI},
  \end{align}
  where $\mathit{f(\cdot)}$ is the cosine similarity function between $\mathcal{F}(x)$ and $\mathcal{R}(z)$.
  $\mathcal{F}(x)$ denotes the pooled textual representation of a source sentence
  $x$ and $\mathcal{R}(z)$ denotes the pooled visual representation from a pre-trained vision model (e.g., ResNet or Vision Transformer). $x^-$ is a negative 
  source sentence drawn from the proposal distribution $q(x^-)$. This contrastive learning objective aligns the textual and visual representations into a unified semantic space, and encourages the text encoder to generate grounded representations, which has shown effectiveness in various scenarios \cite{Elliott2017ImaginationIM,Kdr2017RepresentationOL}.
  
  % \subsection{$I(\mathrm{Y};\mathrm{Z}|\mathrm{X})$}
  \subsection{Target-Specific Conditional Mutual Information}
  \label{sec:target-mi}
  In MMT, $I(\mathrm{Y};\mathrm{Z}|\mathrm{X})$ represents how much information the image $Z$ provides corresponding to the target $Y$ given the source $X$.
  Our goal is to maximize the conditional mutual information as follows:
  \begin{align}
      I(\mathrm{Y};\mathrm{Z}|\mathrm{X})=H(\mathrm{Y}|\mathrm{X})-H(\mathrm{Y}|\mathrm{X},\mathrm{Z}).
  \end{align}
  
  Note that the term $H(\mathrm{Y}|\mathrm{X})$ refers to the conditional probability $P(Y|X)$, which is non-accessible in a MMT model. 
  To remedy this issue, we instead introduce a pseudo conditional mutual information  $I(\mathrm{Y};\widetilde{\mathrm{Z}}|\mathrm{X})$, which is formulated as:
  \begin{align}
      I(\mathrm{Y};\widetilde{\mathrm{Z}}|\mathrm{X})=H(\mathrm{Y}|\mathrm{X})-H(\mathrm{Y}|\mathrm{X},\widetilde{\mathrm{Z}}),
  \end{align}
%   new objective $I(\mathrm{Y};\widetilde{\mathrm{Z}}|\mathrm{X})$.
  % $$I(\mathrm{Y};\widetilde{\mathrm{Z}}|\mathrm{X})=H(\mathrm{Y}|\mathrm{X})-H(\mathrm{Y}|\mathrm{X},\widetilde{\mathrm{Z}}).$$
  where $\widetilde{\mathrm{Z}}$ represents a set of \textbf{deteriorated images} $(\widetilde{z_1},\widetilde{z_2},\cdots,\widetilde{z_n})$. The visual information of these deteriorated images are artificially interfered by some means. For a well-trained visual-aware MMT, we suppose a deteriorated image would not significantly reduce the uncertainty of target sentence prediction compared to its text-only counterpart. Thus our goal is to minimize the pseudo conditional mutual information $I(\mathrm{Y};\widetilde{\mathrm{Z}}|\mathrm{X})$. Accordingly, we have the training objective as follows:
\begin{align}
  I(\mathrm{Y};\mathrm{Z}&|\mathrm{X})-I(\mathrm{Y};\widetilde{\mathrm{Z}}|\mathrm{X}) 
   \nonumber\\
   =& -H(\mathrm{Y}|\mathrm{X}, \mathrm{Z}) + H(Y|\mathrm{X},\widetilde{\mathrm{Z}}) \nonumber\\
   ~=& \sum_{x,y,z}p(x,y,z)\log{p(y|x,z)} \nonumber\\
   ~&- \sum_{x,y,\widetilde{z}}p(x,y,\widetilde{z})\log{p(y|x,\widetilde{z})}.
\end{align}

 Minimizing $I(\mathrm{Y};\widetilde{\mathrm{Z}}|\mathrm{X})$ means that the MMT can correctly discriminate the useless visual information. 
 Moreover, maximizing $I(\mathrm{Y};\mathrm{Z}|\mathrm{X})$ means that the MMT model is able to make the most of the grounding information from images. 
  
 However, the above training objective cannot be directly calculated since the probability mass function $p(x,y,z)$ is unobtainable.
 Therefore, we follow  \citet{Bugliarello2020ItsET,Fernandes2021MeasuringAI} to use a Monte Carlo estimator to approximate it. Under sufficient held-out training data of $\{(x^{i},y^{i},z^{i})\}_{i=1}^N$, we can estimate
  the true cross-entropy as follows:
  \begin{align}
      H(\mathrm{Y}|\mathrm{X}, \mathrm{Z}) \approx -\frac{1}{N} \sum_{i=1}^{N}\log{p(y^{(i)}|x^{(i)},z^{(i)})}.
  \end{align}
  
  This estimation will get more accurate when the training data set is large enough. We note that $\widetilde{z}$ is an deteriorated version of the image $z$ and is also drawn from the same true probability
  distribution. We can employ the same estimation method to approximate the cross-entropy. Finally, we have an estimation for the optimization objective:
  \begin{align}
      \mathcal{L}_{TMI}
      = -\frac{1}{N} \sum_{i=1}^{N} \log{\frac{p(y^{(i)}|x^{(i)},z^{(i)})}{p(y^{(i)}|x^{(i)},\widetilde{z}^{(i)})}}.
  \end{align}
  
  Moreover, the training process can be unstable when the term $p(y^{(i)}|x^{(i)},\widetilde{z}^{(i)})$ becomes very small, ignoring the fact that $x$ has already contained the majority of information for translation. Following \citet{Huang2018LargeMN}, we perform back propagation only when $\log{\frac{p(y^{(i)}|x^{(i)},z^{(i)})}{p(y^{(i)}|x^{(i)},\widetilde{z}^{(i)})}}$ is lower than a margin value $m$. 
  The above objective can be converted into:
  \begin{align}
      \mathcal{L}_{TMI}
      = \frac{1}{N} \sum_{i=1}^{N} \max\{0, m-\log{\frac{p(y^{(i)}|x^{(i)},z^{(i)})}{p(y^{(i)}|x^{(i)},\widetilde{z}^{(i)})}\}
      }.
  \end{align}
  \subsection{Deteriorated Image Generation}
  In this section we describe how to generate a deteriorated view of the image $z$. One straightforward approach is masking relevant objects in the visual modality by using an object detector. This method can reduce the amount of information contained in a picture. However,
%   an object detector can not ensure a high detection accuracy, especially when the image is out-of-domain.
  using an external object detector is time-consuming, making the MMT task more complex.
  In this work, we adopt a more efficient and feasible dropout-based method to generate the deteriorated images at the representation level. Specifically, we randomly select the units in the visual vector and mask them with zeros by following a Bernoulli distribution of probability $\rho$. In this way, the majority of information in the image representations will be corrupted.
%   This practice is similar to dropout but has different purposes: our methods aims to erase the majority of information in the image representations while dropout discourages the network to overfit the training data.
  
  \subsection{Model Architecture}
  We use a simple gated mechanism for multimodal fusion, which has achieved competitive results versus other complicated models. Following the recent work \cite{Wu2021GoodFM}, the source sentences and the images are separately encoded by using a transformer-based encoder and a pre-trained Resnet model. 
  Formally, the textual and image representations are combined by a learnable gate vector $\lambda$ as:
  \begin{align}
      ~&\lambda = \textrm{Sigmoid}(W_{\lambda}\cdot\textrm{Concat}[\mathcal{F}(x);\mathcal{R}(z)]), \nonumber\\
      ~&\mathrm{H_{fusion}} = \mathcal{F}(x) + \lambda  \odot \mathcal{R}(z).
  \end{align}
$W_{\lambda}$ is a trainable matrix and $\odot$ denotes element-wise product. The decoder takes the modality input $\mathrm{H_{fusion}}$ in the recurrent fashion as the text-only NMT does. The deteriorated image input is fused in the same way by replacing $\mathcal{R}(z)$ with $\mathcal{R}(\widetilde{z})$.

  \begin{table*}[htbp]
    \centering
    % \small
    % \renewcommand\arraystretch{1.5}
    \begin{tabular}{cccccc}
      \toprule[1.0pt]
      % \hline
      \multicolumn{1}{c|}{\multirow{2}{*}{Row}} & \multicolumn{1}{c|}{\multirow{2}{*}{Method}} & \multicolumn{2}{c|}{Validation}                           & \multicolumn{2}{c}{Test}            \\ \cline{3-6} 
      \multicolumn{1}{c|}{}                     & \multicolumn{1}{c|}{}                        & \multicolumn{1}{c|}{BLEU} & \multicolumn{1}{c|}{METEOR} & \multicolumn{1}{c|}{BLEU} & METEOR \\
      \hline \hline
      \multicolumn{6}{c}{\textbf{\textit{Only Text}}}                                                           \\ \hline
      \multicolumn{1}{c|}{1}                    & \multicolumn{1}{c|}{Transformer$^\spadesuit$\cite{Li2021VisionMW}}             & \multicolumn{1}{c}{-}  & \multicolumn{1}{c}{-}   & \multicolumn{1}{|c}{35.71}   & -   \\ \hline
      \multicolumn{1}{c|}{2}                    & \multicolumn{1}{c|}{Transformer$^\diamondsuit$}             & \multicolumn{1}{c}{37.46}  & \multicolumn{1}{c}{69.31}   & \multicolumn{1}{|c}{37.60}   & 68.64   \\ \hline
      \multicolumn{6}{c}{ \textbf{\textit{Existing MMT Systems}}}                                                                                                                                                   \\ \hline
      \multicolumn{1}{c|}{3}                    & \multicolumn{1}{c|}{Imagination$^\diamondsuit$\cite{Elliott2017ImaginationIM}}             & \multicolumn{1}{c}{37.83}  & \multicolumn{1}{c}{68.92}   & \multicolumn{1}{|c}{38.11}  & 69.25   \\ 
     \multicolumn{1}{c|}{4}                    & \multicolumn{1}{c|}{Gated Fusion$^\spadesuit$\cite{Li2021VisionMW}}            & \multicolumn{1}{c}{-}      & \multicolumn{1}{c}{-}      & \multicolumn{1}{|c}{36.68}  & -      \\ 
      \multicolumn{1}{c|}{5}                    & \multicolumn{1}{c|}{Selective Attn$^\diamondsuit$\cite{Li2022OnVF}}          & \multicolumn{1}{c}{38.47}       & \multicolumn{1}{c}{68.91}       & \multicolumn{1}{|c}{38.30}       &   69.31     \\ \hline
      \multicolumn{6}{c}{\textbf{\textit{Our MMT System}}}                     \\ \hline
      \multicolumn{1}{c|}{6}                    & \multicolumn{1}{c|}{Gated \quad Fusion$^\diamondsuit$}             & \multicolumn{1}{c}{38.29}  & \multicolumn{1}{c}{69.62}   & \multicolumn{1}{|c}{38.06}  & 69.13   \\ 
      \multicolumn{1}{c|}{7}                    & \multicolumn{1}{c|}{Our \quad Method}                    & \multicolumn{1}{c}{\textbf{39.24}}  & \multicolumn{1}{c}{\textbf{70.30}}   & \multicolumn{1}{|c}{\textbf{39.40}}   & \textbf{70.22}   \\ 
      \bottomrule[1.0pt]
    \end{tabular}
    \caption{Tr $\Rightarrow$ En translation results on AmbigCaps dataset. 
    $\spadesuit$ indicates the results come from the original papers and $\diamondsuit$ indicates the results are from our implementation. All
    models are based on the transformer-tiny configuration.}
    \label{tab:ambigcapsRes}  
    \end{table*}  

  \subsection{Training Objective}
  Finally, the two MI-based objectives are jointly optimized with MMT loss from scratch:
  \begin{align}
      \mathcal{L}
      = \mathcal{L}_{MMT} + \alpha|s|\mathcal{L}_{SMI} + \beta|s|\mathcal{L}_{TMI}, 
  \end{align}
  where $\alpha$ and $\beta$ $\in [0,1]$ are the hyperparameters for balancing MI maximizing objective and MT loss, and $|s|$ denotes the average length of the sequence since the MI-based loss is sentence-level \cite{Pan2021ContrastiveLF}. 
%   Figure \ref{fig:model} illustrates the overview of our proposed method.

\section{Experiments}
  
  \subsection{Data}
We evaluate our methods on two standard MMT datasets, including \textbf{AmbigCaps}
\cite{Li2021VisionMW} 
and \textbf{Multi30K} \cite{Elliott2016Multi30KME}.

AmbigCaps contains 81K Turkish-English parallel sentence pairs with visual annotations. We use 1,000 sentences provided by Li et al. \shortcite{Li2021VisionMW} for validation and testing, respectively. This dataset is processed into a  gender-ambiguous one, in which sentences containing the gender of the entity and professions with gender implications removed. Moreover, AmbigCaps is designed to translate a gender-neutral language (Turkish) into a gender-specified language (English), which is suitable for the MMT task since the source
textual information is not naturally sufficient.
  
Multi30K is a widely used dataset for MMT, which contains 29K images with one English description and a German manual translation.  We follow a standard split for experiments and use 1,014 sentences for validation and 1,000  sentences for testing (Test2016). In addition, we also evaluate the WMT17 test set (Test2017) and the ambiguous MSCOCO test set, which include 1,000 and 461 triplets, respectively.

\begin{table*}[t]
  \centering
  \begin{tabular}{c|cc|cc|cc}
  \toprule[1.5pt]
  \multirow{2}{*}{Method} & \multicolumn{2}{c}{Test2016}      & \multicolumn{2}{c}{Test2017}      & \multicolumn{2}{c}{MSCOCO}        \\ \cline{2-7} 
      & \multicolumn{1}{c}{B}   & M     & \multicolumn{1}{c}{B}   & M     & \multicolumn{1}{c}{B}   & M     \\ 
    \midrule[1pt]
  Transformer \cite{Vaswani2017AttentionIA}            & \multicolumn{1}{c}{41.02} & 68.22 & \multicolumn{1}{c}{33.36} & 62.05 & \multicolumn{1}{c}{29.88} & 56.64 \\ 
  Imagination  \cite{Elliott2017ImaginationIM}           & \multicolumn{1}{c}{41.31} & 68.06 & \multicolumn{1}{c}{32.89} & 61.29 & \multicolumn{1}{c}{29.9}  & 56.57 \\ 
  UVR  \cite{Zhang2020NeuralMT}                   & \multicolumn{1}{c}{40.79} & - & \multicolumn{1}{c}{32.16} & -     & \multicolumn{1}{c}{29.02} & -     \\ 
  DCCN  \cite{Lin2020DynamicCC}                  & \multicolumn{1}{c}{39.7}  & -     & \multicolumn{1}{c}{31.0}    & -     & \multicolumn{1}{c}{26.7}  & -     \\ 
  Multimodal Graph\cite{Yin2020ANG}        & \multicolumn{1}{c}{39.8}  & -     & \multicolumn{1}{c}{32.2}  & -     & 
  \multicolumn{1}{c}{28.7}  & -     \\ 
      Gated Fusion  \cite{Wu2021GoodFM}          & \multicolumn{1}{c}{41.55} & 68.34 & \multicolumn{1}{c}{33.59} & 61.94 & \multicolumn{1}{c}{29.04} & 56.15 \\ 
    Selective Attn \cite{Li2022OnVF}         & \multicolumn{1}{c}{\textbf{41.93}} & 68.55 & \multicolumn{1}{c}{33.62} & 61.61 & \multicolumn{1}{c}{29.72} & \textbf{56.94} \\ \midrule[1pt]
  Our \quad Method               & \multicolumn{1}{c}{41.77} & \textbf{68.60}  & \multicolumn{1}{c}{\textbf{34.58}} & \textbf{62.40}  & \multicolumn{1}{c}{\textbf{30.61}} & 56.70  \\
  \bottomrule[1.5pt]
\end{tabular}
\caption{En $\Rightarrow$ De results on Multi30K dataset. Some results are from Wu et al.  \shortcite{Wu2021GoodFM} and Li et al. \shortcite{Li2022OnVF}. B and M denote BLEU and METEOR score.}
\label{tab:multi30kRes}
\end{table*}

\subsection{System Setting}
We use the Transformer-Tiny \cite{Vaswani2017AttentionIA} configuration to conduct all of our experiments, which can even obtain better performance than those large models \cite{Wu2021GoodFM}. The tiny transformer model consists of 4 encoder
and decoder layers, with 1024 embedding/hidden units, 4096 feed-forward filter
size and 4 heads per layer. We employ the Adam optimizer with $\text{lr}=0.005$, $t_{\text{warm}\_\text{up}}=2000$ and $\text{dropout}=0.3$ for optimization. At the training time, each training batch includes 4096 source tokens and target tokens. For a fair comparison with previous works, we use the early stopping strategy if the performance on validation does not gain improvements for ten epochs on Multi30K.
Empirically, $1e^{-3}$ for $\alpha$ and $1e^{-4}$ for $\beta$ can get the best results. 
The model for AmbigCaps is trained on a single 
Tesla P40 GPU and the one for Multi30k is trained on two GPUs for a fair comparison with previous work.

In particular, the number of patience on AmbigCaps is set to twenty epochs since the converge speed on this dataset is slow. We average the results of the last ten checkpoints during inference. At decoding time, we generate with $beam\_size=5$ and length penalty is equal to $1.0$.

For visual features, we use a ResNet-50 \cite{He2016DeepRL} model pre-trained on ImageNet as the image encoder. The dimension of the global feature is 2048 and we use a projection matrix to convert the shape of image features into that of text features. For data processing, we generate subwords following Sennrich et al. \shortcite{Sennrich2016NeuralMT} with 10,000 merging operations jointly
to segment words into subwords, which generates a vocabulary of  6,765 and 9,712  tokens for AmbigCaps and Multi30K, respectively.

Finally, we report BLEU\footnote{\small{\url{https://github.com/moses-smt/mosesdecoder/blob/master/scripts/generic/multi-bleu.perl}}} \cite{Papineni2002BleuAM} and METEOR\footnote{\small{\url{https://github.com/facebookresearch/vizseq/blob/main/vizseq/scorers/meteor.py}}} \cite{Denkowski2014MeteorUL} scores to evaluate the 
quality of different translation system. Our implementation are implemented on Fairseq \cite{Ott2019fairseqAF}.

\subsection{Baseline Methods}
We compare our method with baselines as follows:
\begin{itemize}
  \item \textbf{Vanilla Transformer} \cite{Vaswani2017AttentionIA}. We report the Transformer-tiny results.
  \item  \textbf{Imagination} \cite{Elliott2017ImaginationIM,Helcl2018CUNISF}. This method adopts a margin loss as a regularizer to the text encoder in order to learn grounded representations. We  reimplement this method in our Transformer-based models instead of RNNs for a fair comparison.
  \item  \textbf{Gated Fusion} \cite{Wu2021GoodFM}.  It  uses a gate vector  to combine textual representations and image representations and then feds them into the MMT decoder. This is a simple yet effective method and can outperform most of MMT systems. 
  \item \textbf{Selective Attn} \cite{Li2022OnVF}. This method has the same model architecture with \textbf{Gated Fusion}
  and supports more fine-grained features extracted from images.
\end{itemize}

Since \textbf{AmbigCaps} is a new  MMT dataset, we reproduce the above baseline methods and report the corresponding results. 

% \begin{table}[t]
% \small
%   \centering
%   % \renewcommand\arraystretch{1.5}
% %   \small
%   \begin{tabular}{c|c|c|c}
%     \toprule[1.2pt]
%     &  ID $\uparrow$ & IS $\downarrow$ & GA $\uparrow $\\ 
% \midrule[1pt]
%   Gated Fusion & 3.32                 & 0.91              & 79.73\%         \\ 
%   Our Method  & 8.31                 & 0.75              & 80.65\%         \\ \bottomrule[1.2pt]
% \end{tabular}
% \caption{visual awareness evaluation from different  methods. $\uparrow$ indicates the awareness is better as the value is larger, while $\downarrow$ otherwise.} 
% \label{ta:adv}
% \end{table}

% \subsection{Results}
\subsection{Results on Tr$\Rightarrow$En  Translation Task}
The evaluation results of different MMT systems on the  Tr $\Rightarrow$ En translation task are presented in Table \ref{tab:ambigcapsRes}.
By observing the reported results, we draw the following interesting conclusions:

{\bf First}, we observe that our implementation for the text-only transformer and Gated Fusion outperform the results reported in the original paper. This is due to the larger number of patience. It also indicates our method is more competitive and still achieves improvement against the strong baselines.

{\bf Second}, all of the MMT models beat the text-only transformer, which indicates that AmbigCaps is a suitable dataset and verifies the benefit of the visual modality. Gated Fusion gains 0.83 BLEU points and 0.31 METEOR points compared to the text-only model. Selective Attn shows further improvement due to the more-grained features. Our model significantly outperforms the text-only transformer by 1.8 on BLEU and 1.58 on METEOR. 

{\bf Third}, our model also beats Imagination, Selective Attn and Gated Fusion. The gap between different baseline results is not significant. Note that the Imagination model does not use the image during the inference time.
% Selective Attn employs fine-grained features which can preserve
% more detailed information than a single global vector.
Our model uses the same model architecture with Selective Attn and Gated Fusion and consistently improves the BLEU score, 1.30 and 1.34 points more than Selective Attn and Gated Fusion. It indicates that our method can significantly utilize the image context and improve translation by using the same model configuration.

\subsection{Results on En$\Rightarrow$De  Translation Task}
Table \ref{tab:multi30kRes} shows the main results on the En$\Rightarrow$De translation task. We find that the text-only transformer can act as a solid baseline with the aid of the elaborate model configuration and training settings, which even beats DCCN and Multimodal Graph models. The BLEU score of our method on Test2016 is quite marginal compared to the base model Gated Fusion, which is the point that we attempt to improve. We argue that the Test2016 test set may contain less textual ambiguity. Besides, we observe significant BLEU/METEOR gains on Test2017 and MSCOCO, outperforming the text-only baseline by 1.22/0.73 points and 0.99/1.57 points. Moreover, our method beats all robust baseline systems on the two test sets.  
These results further indicate the necessity of visual context for multimodal translation.

\section{Model Analysis}

% \begin{table}[t]
% \small
%   \centering
% %   \small
% %   \renewcommand\arraystretch{1.5}
%   \begin{tabular}{l|c|c|c|c|c}
%     \toprule[1.2pt]
%     System                        & B@4  & M   &   ID $\uparrow$ & IS $\downarrow$ & GA $\uparrow $ \\ 
%     \midrule[1pt]
%     $\mathcal{L}_{MMT}$            & 38.06 &   69.13    & 3.32 & 0.91 & 79.73\% \\ 
%     \multicolumn{1}{r|}{+ $\mathcal{L}_{SMI}$} & 38.72 &   69.50  & 3.35 & 0.39 & 80.23 \%     \\ 
%     \multicolumn{1}{r|}{+ $\mathcal{L}_{TMI}$} & 38.74 &   69.93 & 4.37 & 0.67 & 79.93\%  \\ 
%     Full Model            & 39.40 &   70.22 & 8.31 &    0.75 & 80.65\%  \\  \bottomrule[1.2pt]
%   \end{tabular}
%     \caption{Ablation study on AmbigCaps. B@4 and M denote BLEU and METEOR score.  $\uparrow$ indicates the visual awareness is better as the value is larger, while $\downarrow$ otherwise.}
%     \label{tab:ablation}
%   \end{table}

\begin{table}[t]
  \centering
%   \small
%   \renewcommand\arraystretch{1.5}
  \begin{tabular}{l|c|c|c|c}
    \toprule[1.2pt]
    System                        & B  & M   &   ID $\uparrow$ & GA $\uparrow $ \\ 
    \midrule[1pt]
    $\mathcal{L}_{MMT}$            & 38.06 &   69.13    & 3.32 & 79.73\% \\ 
    \multicolumn{1}{r|}{+ $\mathcal{L}_{SMI}$} & 38.72 &   69.50  & 3.35  & 80.23 \%     \\ 
    \multicolumn{1}{r|}{+ $\mathcal{L}_{TMI}$} & 38.74 &   69.93 & 4.37  & 79.93\%  \\ 
    Full Model            & 39.40 &   70.22 & 8.31 & 80.65\%  \\  \bottomrule[1.2pt]
  \end{tabular}
    \caption{Ablation study on AmbigCaps. B and M denote BLEU and METEOR score.  $\uparrow$ indicates the visual awareness is better as the value is larger.}
    \label{tab:ablation}
  \end{table}

\subsection{Ablation Study}

To make a better understanding of the influence of the training objective, we perform ablation studies to validate the impact of each part on translation quality. 
Column 2 to column 3 in Table \ref{tab:ablation} shows both $\mathcal{L}_{SMI}$ and $\mathcal{L}_{TMI}$ results in a close gain in BLEU score, 0.66 and 0.68 points more than the vanilla MMT model. We find $\mathcal{L}_{TMI}$ can perform better than $\mathcal{L}_{SMI}$ in METEOR score. Combining the two training objectives can further boost the translation performance, which means $\mathcal{L}_{SMI}$ and $\mathcal{L}_{TMI}$  can be beneficial to each other. It indicates that our method improves the image utilization at both the source and target sides.

Besides, we also investigate how the visual signals help to translate and whether the gains on translation come from the increasing visual awareness . To achieve this goal, we introduce two new
metrics to evaluate the visual awareness of the MMT model, which are \textit{Incongruent Decoding} and  \textit{Gender Accuracy.}  We will first explain how these metrics are calculated.

{\bf Incongruent Decoding (ID)}\ \ We follow Elliott et al. \shortcite{Elliott2018AdversarialEO}
to use an adversarial evaluation method to test if our method is more sensitive to the visual context. We randomly replace the image with an incongruent image or set the congruent image's feature to all zeros. Then we observe the value $\bigtriangleup \rm BLEU$ by calculating the difference between the congruent data and the incongruent one. A larger value means the model is more sensitive to the image context. 

% {\bf Incomplete Source (IS)} \ \ We also consider how the image context helps when the source text is incomplete. We degrade the textual inputs by randomly replacing the tokens in source texts with the \textit{UNK} token. The $\bigtriangleup \rm BLEU$ is still used for evaluation. On the contrary, a slight fluctuation indicates the model is more visual-aware since the MMT model can gain visual information even if the text is incomplete.

{\bf Gender Accuracy (GA)}\ \ Following Li et al. \shortcite{Li2021VisionMW}, we divide the translation sentences into three categories:
\textit{male, female} and \textit{undetermined}. \textit{Male} means that the target sentences contains at least one of the male pronouns \{‘he’, ‘him’, ‘his’, ‘himself’\}.  
The definition of \textit{Female} is similar to \textit{Male}. The sentence is \textit{undetermined} when it contains both male and female pronouns or neither, and will not be considered for the calculation of gender accuracy. It is designed to determine whether the ambiguous words are indeed translated correctly.

Column 4 to column 5 in Table \ref{tab:ablation} show the results of visual awareness evaluation. 
When given the incongruent image, $\mathcal{L}_{MMT}$ drops more than 3 points, indicating that this simple method can also utilize the visual context to some extent.  When   $\mathcal{L}_{SMI}$ 
or $\mathcal{L}_{TMI}$ is used, the dropped value  will become larger. Combining the two objective 
can lead to a great decline. It means the visual part plays a vital role in our system.  There is also a significant increase in gender accuracy with our method. 
It also proves that the translation improvement partly benefits from the correct translation of ambiguous words. 

  \begin{figure}[htbp]

\centering
\includegraphics[width=0.73\linewidth]{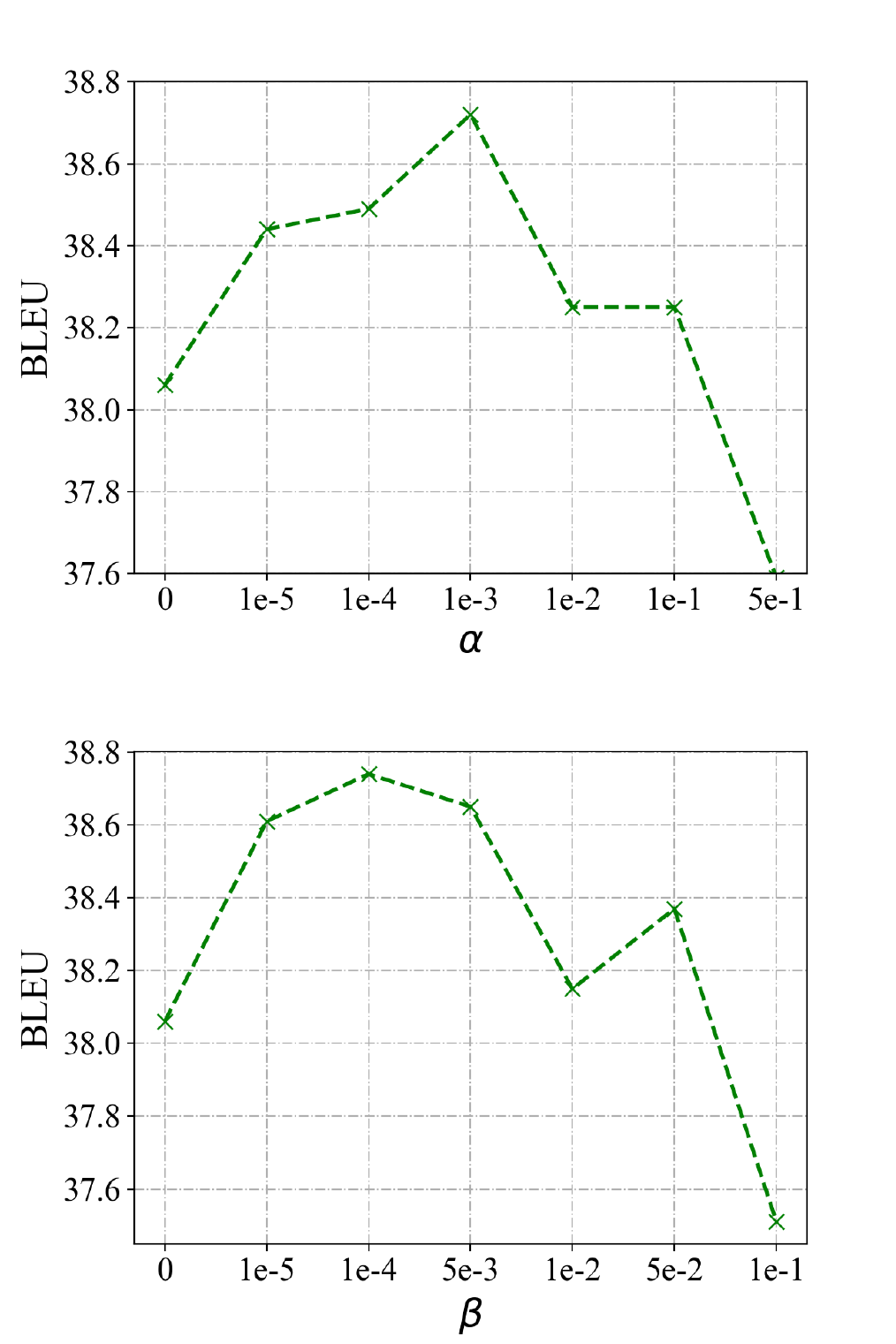}
\caption{Illustration of BLEU score  with different $\alpha$ (left) and $\beta$ (right).}
\label{fig:mi}
\end{figure}

\subsection{The Influence of Weight $\alpha$ and $\beta$}
To evaluate the effects of the coefficients of $\alpha$ and $\beta$,
we observe the translation performance over various values by setting the corresponding coefficient to zero.
Figures \ref{fig:mi} shows the performance fluctuates greatly with a low rate of 
$\alpha$ and $\beta$.
With the coefficients increasing slightly, the BLEU score improves, which probes the effectiveness of increasing visual awareness by applying our method.
The translation performance drops slightly 
as the coefficients increase since the MI-based loss will dominate the optimization and ignore the MLE training objective.
In particular, the best performance is achieved when $\alpha=1e^{-3}$ or $\beta=1e^{-4}$ respectively.

\begin{figure}[htbp]
  \centering % 图片居中
  \includegraphics[width = 0.8 \linewidth]{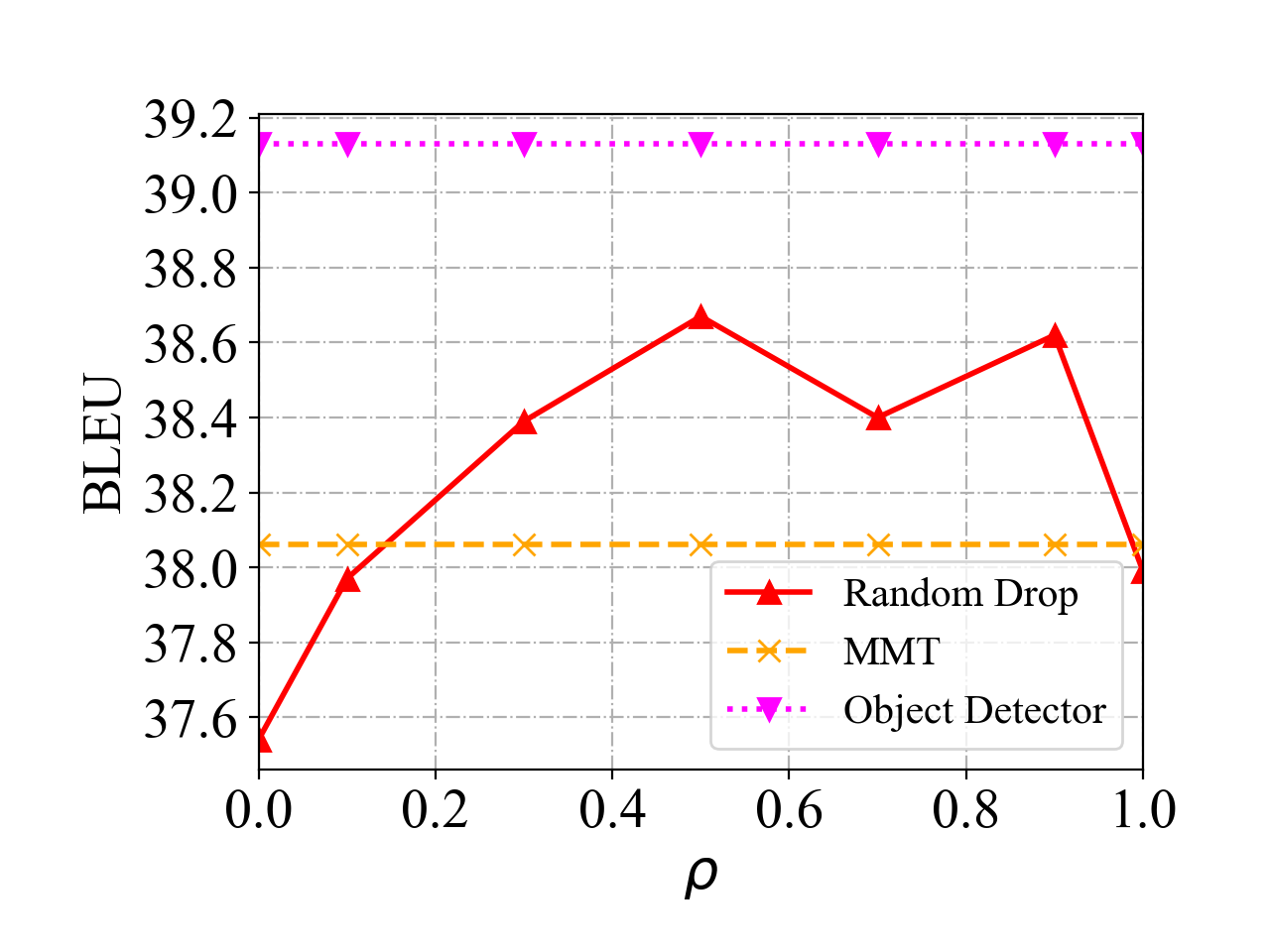}
 \caption{
    Illustration of BLEU score on AmbigCaps with various methods.
%   We fix $\beta=1e-4, m=0.3$. $\alpha$ is set to zero for excluding the effect of $\mathcal{L}_{SMI}$.
  The orange dash line denotes the model that only uses $\mathcal{L}_{MMT}$.
  The red solid line denotes the one that masks features with different probabilities $\rho$. The fuchsia dotted line denotes the one that uses an object detector for masking.
  }
  \label{fig:dprate}
\end{figure}

\subsection{The Influence of Deteriorated Image Generation.}
First, we compare the effects of
different methods to generate the deteriorated image in Figure \ref{fig:dprate}. 
We find that using a object detector for masking  relevant objects can generate the best translation results. It is not surprising since 
the picture in AmbigCaps is not complex and the detector model is competent to detect the most objects correctly. We should note that our method is 
a unsupervised manner but can still generate the competitive results.
We also explore the different influence of $\mathcal{L}_{TMI}$
with the various dropout rate $\rho$. 
When the value of $\rho$ is very small, the BLEU score is lower than the baseline. This is due to that a small value of $\rho$ means that most image information is preserved and the term $I(\mathrm{Y};\widetilde{\mathrm{Z}}|\mathrm{X})$  should be as large as 
possible instead of being minimized.
Meanwhile, $\rho=1$  means that the image is totally unused, and the global vector is all zeros.
This conspicuous feature induces the model to take a shortcut to discriminate the useless image easily.
Only the appropriate value of $\rho\geq 0.5$ can get stable and good results. 
It indicates a well-designed  method to generate deteriorated image is critical for better translation.

\section{Related Work}
\subsection{Mutual Information Maximization} Mutual information has been 
widely explored in recent NLP work. Bugliarello et al. \shortcite{Bugliarello2020ItsET} proposed 
cross-mutual information to measure the neural translation difficulty.
Fernandes et al. \shortcite{Fernandes2021MeasuringAI}
used conditional mutual information as the metric to estimate the context usage 
in document NMT. Our work differs in that we employ MI as a training objective to
enhance image usage instead of a pure metric. Wang et al. \shortcite{Wang2022MINERIO}
maximized MI to solve the OOV problem in NER. Kong et al. \shortcite{Kong2020AMI} improved language model pre-training from a  mutual information maximization perspective.
To our best knowledge, we are the first to employ mutual information to increase image awareness in MMT generation task.

\subsection{Efficient Visual Context  Modeling}
To equip the MMT model with the visual context modeling ability,
Caglayan et al. \shortcite{Caglayan2021CrosslingualVP} and Song et al. \shortcite{Song2021ProductorientedMT}
used cross-modal pre-training to learn visually-grounded representations.
This point is similar to our source-related MI method, which aligns 
the textual and visual representations in a shared semantic space.
Wang and Xiong \shortcite{Wang2021EfficientOV}  
forced the model to reward masking irrelevant objects and penalize
masking relevant objects. It also makes the model sensitive to the changes in 
the visual modality and enhances visual awareness. But detecting the relevant objects and correlating the regions with the vision-consistent target words are rather complex.
\subsection{Contrastive Learning}
Our method is highly related to contrastive learning. The source-specific loss is optimized 
by using the $\textbf{InfoNCE}$  bound, and the target-specific loss can also be viewed as 
using a contrastive image form and lowering the probability of generating the target.
Contrastive learning has been recently used in translation tasks.
Pan et al. \shortcite{Pan2021ContrastiveLF} closed the gap among representations of different languages
to improve the multilingual translation.
Zhang et al. \shortcite{Zhang2021FrequencyAwareCL} meliorated word representation space to solve the low-frequency word prediction.
Hwang et al. \shortcite{Hwang2021ContrastiveLF} used coreference information for generating the contrastive samples to train the model to be sensitive to coreference inconsistency. 
Different from this work, our method does not need external knowledge.

\section{Conclusion}
In this paper, we summarize that only using the MLE objective is unable to ensure the image information
is fully utilized, and there still exists tremendous potential to explore.
We propose a method based on mutual information to increase visual awareness in multimodal machine translation.
The information from images is divided into two parts(source-specific and target-specific).
We quantify this contribution from the image and put forward the corresponding optimization solutions. 
Experiments on Multi30K and AmbigCaps show the superiority of our methods.
Detailed analysis experiments conducted probes that our methods are quite interpretable.
In the feature, we would like to apply our MI-based method to visual language pretraining
and other unsupervised tasks to improve the downstream tasks.

\section{Limitations}
Since we conduct all experiments on the machine translation task, it is unclear  whether our approach can benefit other multimodal tasks, e.g. Visual QA  or multimodal Dialog.
Another thing worth noting is that the image-source-target triplet data is not easily available in reality. Thus how to effectively use the unpaired  image in the unsupervised way is a more  promising research.

\bibliography{mybibliography}
\bibliographystyle{acl_natbib}

\end{document}